\documentclass[conference]{IEEEtran}
\IEEEoverridecommandlockouts
\usepackage{balance}
\usepackage{amsmath,amssymb,amsfonts}
\usepackage{algorithmic}
\usepackage{graphicx}
\usepackage{textcomp}
\usepackage{xcolor}
\usepackage{comment}
\usepackage[export]{adjustbox}
\usepackage{multirow}
\usepackage{amssymb}
\usepackage{mathptmx}
\usepackage{graphicx}
\usepackage{comment}
\usepackage{xspace}
\usepackage{tabularx}
\usepackage{float}
\usepackage{algorithm}
\usepackage{algorithmic}
\usepackage{mathtools}
\usepackage{amsmath}
\usepackage{hyperref}
\usepackage{caption}
\usepackage{subcaption}
\usepackage{booktabs}
\usepackage{multirow}
\hypersetup{
	colorlinks=true,
	linkcolor=blue,
	filecolor=mdecision makera,      
	urlcolor=cyan,
}

\def\BibTeX{{\rm B\kern-.05em{\sc i\kern-.025em b}\kern-.08em
    T\kern-.1667em\lower.7ex\hbox{E}\kern-.125emX}}
\begin{document}

\title{Dynamic Decision Making in Engineering System Design: A Deep Q-Learning Approach
}

\author{\IEEEauthorblockN{Ramin Giahi, Cameron A. MacKenzie, Reyhaneh Bijari}

\IEEEauthorblockA{\textit{Department of Industrial and Manufacturing Systems Engineering, Iowa State University, Ames, IA 50011 USA} \\
\{giahi.ramin@gmail.com; camacken@iastate.edu; rbijari@iastate.edu\}}
}


\maketitle

\begin{abstract}

Engineering system design, viewed as a decision-making process, faces challenges due to complexity and uncertainty. In this paper, we present a framework proposing the use of the Deep Q-learning algorithm to optimize the design of engineering systems. We outline a step-by-step framework for optimizing engineering system designs. The goal is to find policies that maximize the output of a simulation model given multiple sources of uncertainties. The proposed algorithm handles linear and non-linear multi-stage stochastic problems, where decision variables are discrete, and the objective function and constraints are assessed via a Monte Carlo simulation. We demonstrate the effectiveness of our proposed framework by solving two engineering system design problems in the presence of multiple uncertainties, such as price and demand.

\end{abstract}



\section{Introduction}
\label{introduction}

Designing engineered systems, such as energy and transportation systems, presents significant challenges, especially considering their prolonged operation under diverse conditions. 
The decision-making process becomes difficult due to the inherent future uncertainties during the design phase \cite{cardin2017approach}. Designers must rely on the best available information, incorporating provisions for future uncertainties. For example, when designing power plants, it is imperative for engineers to account for uncertainties associated with both the price and demand for electricity. Furthermore, it may become crucial to incorporate flexibility into the initial design, enabling adaptability to potential shifts in factors such as electricity prices, demand dynamics, and technological advancements for renewable energy sources that may unfold over time. This dynamic nature, where initial designs may evolve over time due to changing conditions, defines engineering system design as a decision-making process characterized by complexity and uncertainty \cite{zhang2011new}.  

When foreseeing future modifications, capacity additions, or system alterations, the design decision becomes a multi-stage problem often impacted by stochastic elements due to uncertain future conditions. To address this, designers commonly tackle multi-stage stochastic optimization problems to determine the optimal design and its modifications under varying conditions \cite{cardin2014enabling, giahi2022optimizing}. The initial design may periodically change in the future based on the evolution of the system. Neglecting to account for uncertainties may lead to substantial future costs in modifying the initial design. Sequential decision problems occur in a range of engineering real-world problems such as robot control \cite{kent2018learning,mihaylova2002comparison,parker2009cooperative}, power plant design \cite{carazas2010risk,meigounpoory2008optimization}, and supply chain management \cite{fattahi2018multi,shabani2016hybrid,nickel2012multi}. Real-world problems involve sequential decision-making under unknown parameters at the decision points \cite{shapiro2007tutorial}.

Evaluating engineering system design often requires employing large-scale and computationally expensive simulation models. Complex simulations are crucial in assessing the performance and accounting for multiple sources of uncertainty \cite{shabani2016evaluating} in designing problems such as designing power plants \cite{amin2011smart}, architectural design \cite{goldstein2017simulation}, and mechanical systems \cite{goldstein2018multiscale, berquist2017investigation}.

Computing the optimal policy for a system, wherein the objective function is assessed through simulation in sequential settings, is challenging \cite{frazier2010decision}. Using dynamic programming to find the global optimal solution to this type of problem is difficult. The curse of dimensionality will arise, and addressing the problem in the context of continuous state spaces, particularly when the objective function is evaluated through simulation, is likely to necessitate the application of heuristics and/or scenario reduction approaches. Solving the exact backward dynamic programming problem requires the evaluation of nearly all non-optimal solutions at time $t$ from the perspective of time $t-1$. For example, if the problem contains 4 stages with a thousand possible realizations of uncertainty, the decision maker would need to solve $1000^{4}$ optimization problems if forward approximate dynamic programming is used \cite{shapiro2005complexity}. 

In contrast, learning algorithms like reinforcement learning mitigate the need for decision-makers to solve an extensive number of optimization problems. Reinforcement learning optimizes system control to maximize a numerical value, reflecting a long-term goal \cite{szepesvari2010algorithms}. In reinforcement learning, an agent engages with and explores the environment, gradually developing an optimal policy that to make the best decision \cite{mnih2015human}.

Reinforcement learning has demonstrated success in addressing diverse challenging problems, including robotics \cite{lillicrap2015continuous}, playing Go \cite{silver2016mastering}, and competitive video games \cite{hester2017deep, gu2016continuous}. Algorithms such as Deep Q-learning \cite{mnih2015human} and Deep Deterministic Policy Gradient \cite{lillicrap2015continuous} have excelled in solving sequential decision-making problems with high dimensional inputs, exemplified by Atari games \cite{lyu2019knowledge}. Reinforcement learning aims to acquire effective policies for sequential decision problems by optimizing cumulative future rewards \cite{van2016Deep}. This paper incorporates a Deep Q-learning algorithm to address a multi-stage stochastic problem in an engineering system where the objective function and constraints are evaluated with the Monte Carlo simulation. The decision-maker interacts with and learns from a stochastic and uncertain environment to find the best policy to optimize the objective function (e.g., profit or cost). 

The primary objective of this paper is to propose a framework for determining the optimal decision in an engineering system when the objective function is evaluated through Monte Carlo simulations. Through reinforcement learning, a decision-maker can identify the optimal policy for a multi-stage stochastic problem without the need to solve numerous optimization problems. Instead, the optimal policy is learned by running simulations and evaluating the objective function. This framework is able to accommodate various simulation models with different types of uncertainty, eliminating the need for assumptions about the distribution of uncertain parameters.

The uniqueness of our proposed framework lies in its utilization of Deep Q-learning to identify the optimal design of an engineering system, where system performance is evaluated through a Monte Carlo simulation, and multiple sources of uncertainty impact system performance and desirability. The methodology presented learns the optimal policy by interacting with the environment via simulation, instead of solving many optimization problems.

\section{Literature Review}
\label{lit}

The field of stochastic programming encompasses various methods aimed at addressing multi-stage stochastic problems. Scenario-based techniques, including scenario generation and reduction, are widely used in stochastic optimization \cite{dupavcova2003scenario}. These approaches approximate the underlying probability distribution of uncertain parameters. As the number of scenarios increases, heuristic methods may be necessary to reduce the computational complexity and time limitations. This simplifies the approximation of distributions using a model with a reduced set of scenarios \cite{romisch2009scenario,growe2003scenario,heitsch2009scenario}. Growe-Kuska et al. \cite{growe2003scenario} developed algorithms for reducing the number of scenarios and creating scenario trees that closely mimic the random data processes in multi-stage dynamic decision models in the electricity power sector. Similarly, Hu and Li \cite{hu2019new} introduced an algorithm centered around the idea of correlation loss, aiming to decrease the number of scenarios. Their method focuses on two main goals: minimizing the correlation loss before and after reducing scenarios and maximizing the similarity between the original and reduced scenario sets. However, a primary limitation of these scenario reduction techniques is the reliance on user-defined scenario numbers, leading to approximate scenario sets that might not cover all potential combinations of uncertain parameters  \cite{wang2016scenario}. 

Decomposition methods are another approach employed in solving multi-stage stochastic problems, particularly useful for large-scale stochastic optimization challenges. These methods break down the main problem into smaller, tractable sub-problems by relaxing coupling constraints \cite{vigerske2013decomposition}. Nested Benders decomposition, a prominent technique in this domain, divides the problem into smaller sections that can be solved separately and then combined to address the main problem \cite{murphy2013benders}. This algorithm has seen extensive application across various fields including power network and energy systems  \cite{soares2017two,gebreslassie2012design}, manufacturing \cite{chu2013integration}, and supply chain  \cite{keyvanshokooh2016hybrid,santoso2005stochastic}. 
Rebennack \cite{rebennack2016combining} combined stochastic dual dynamic programming and the scenario tree framework to incorporate different types of uncertainties. Sherali and
Fraticelli \cite{sherali2002modification} modified the Benders decomposition method using a Reformulation-Linearization Technique along with lift-and-project cuts to tackle discrete optimization problems. Boland et al. \cite{boland2016proximity} developed Proximity Search Benders decomposition, which prioritizes high-quality solutions, but not necessarily optimal solutions. 

Various random search techniques are also developed for solving multi-stage stochastic problems, including pure adaptive search \cite{patel1989pure} and accelerated random search \cite{ombach2014short, appel2004accelerated}. Evolutionary algorithms like genetics algorithm, differential evolution, and simulated annealing have been applied in engineering to address multi-stage stochastic programming challenges \cite{dutta2017genetic,zohrevand2016multi,zheng2019two}.

Numerous studies address the challenge of designing engineering systems under evolving uncertainties. Zhang and Babovic \cite{zhang2011evolutionary} introduced a simulation-based real options valuation for engineering system design, identifying optimal and flexible designs through a discrete and limited set of scenarios in their evolutionary algorithm. Marques et al. \cite{marques2015using} identified the optimal design of a water distribution network considering eight future scenarios. Power and Reid \cite{power2018decision} applied a real options decision framework with regression models to enhance the performance of systems. Cheah and Liu \cite{cheah2006valuing} utilized Monte Carlo simulations in a discounted cash flow model to design flexible infrastructure project designs. Mak and Shen \cite{mak2009stochastic} applied a two-stage integer stochastic programming formulation to optimize the design of a manufacturing system under demand uncertainty. In the first stage, the demand for each product is known; however, in the second it is assumed that the demand follows a probability distribution. The optimal design is found considering twenty products and a small number of scenarios for demand uncertainty.


Simulation-optimization is frequently employed in engineering design when the objective function and/or constraints lack a closed-form formula. Simulation-optimization is used to find the best solution for problems where the objective function and constraints are evaluated through simulation \cite{amaran2016simulation}. For example, Damodaram and Zechma \cite{damodaram2013simulation} used simulation-optimization to identify and explore watershed management plans. Fong et al \cite{fong2011simulation} utilized simulation-optimization to design solar-thermal refrigeration systems. They employed Monte Carlo simulation method to formulate the objective function. MacKenzie and
Hu \cite{mackenzie2019decision} developed a simulation-optimization framework to maximize expected profit for designing resilience in engineered systems. Giahi et al. \cite{giahi2020design} employed a Bayesian optimization algorithm to optimize engineering system designs for risk-averse decision-makers.  Li et al. \cite{li2019high} introduced a framework for optimizing engineered system designs, aiming to minimize design costs while ensuring reliability. This method is extensively applied in the literature for addressing static engineering design issues. Nonetheless, many design problems are dynamic with uncertainties evolving over time, requiring solutions for multi-stage dynamic stochastic problems. However, reinforcement learning method aims to maximize cumulative future rewards. The goal is to learn an optimal policy for maximizing cumulative returns. 

The Deep Q-learning algorithm can avoid the curse of dimensionality that appears in stochastic dynamic programming. The agent learns to make stage-by-stage decisions based on environment conditions and simulation results. The proposed framework can benefit decision-making in engineering system design, accommodating various uncertainty scenarios with objective functions and constraints evaluated via Monte Carlo simulation. The proposed framework can solve a dynamic decision problem where a sequence of decisions is made over time.

\section{Decision-Making Framework}	\label{DecisionMakingFramework}

Multi-stage stochastic programs are a well-recognized model for sequential decision-making conditioned on information revealed at various points in time, referred to as stages \cite{pantuso2019number}. The structure of a linear multi-stage stochastic program is as follows:

\begin{equation}\label{eq:linstch}
\begin{aligned}
 \max f(x, \xi) & = c_{1}(\xi_{1})x_1 + E{ \sum_{t=2}^{T} c_{t}(\xi_{t}) x_{t}} \\
 \text{subject to } & \sum_{\tau=1}^{t} A_{t \tau}\left(\xi_{t}\right) x_{\tau} \le b_{t}\left(\xi_{t}\right), t=1 \ldots, T \\
& x_{t} \in X_{t}, t=1 \ldots, T 
\end{aligned}
\end{equation}

\noindent where $t=1,..., T$ are stages, $\xi := (\xi_1, . . . , \xi_T )$ is a stochastic process defined on some probability space, and $x := (x_1, . . . , x_T )$ is the collection of decisions. The function $f(x, \xi)$ is a linear objective function for the planning time $T$, $c_{t}(\xi_{t})$ is the per-unit reward at time $t$, $A_{t\tau}(\xi_{t})$ is a vector of coefficients, and $b_{t}(\xi_{t})$ is a constraint at time $t$. The stochastic problem in (\ref{eq:linstch}) requires that decisions $x_t$ is a function of the history of decisions until $t-1$, and that alternatives are members of the feasibility set $X_t$. In many practical applications, $\xi$ is often discrete, possibly based on assumption \cite{pantuso2019number}. This assumption presents challenges as stated in \cite{shapiro2005complexity}. The accurate estimation of probability distributions may not be calculated, or they may vary over time, and evaluating the expected value function necessitates the computation of multivariate integrals.  A finite discretization of the random data enables the solution to be calculated in the form of summation.

Solving multi-stage linear stochastic programming problems becomes highly challenging when dealing with a large number of stages and scenarios. 
This problem is even more difficult when the functions are non-linear. The functions are non-linear for many engineering design problems that rely on simulation. 
The non-linear problem introduced in (\ref{eq:blkstch}) can be used to formulate a multi-stage stochastic problem when the objective function is evaluated via a simulation. 

\begin{equation}\label{eq:blkstch}
\begin{aligned}
\max f(x, \xi)  & = g(x_1,\xi_1) + E{ \sum_{t=2}^{T} g(\xi_{t}, x_{t}) } \\
\text{subject to }& h(\xi_{t}, x_{t}) \le b_{t}\left(\xi_{t}\right), t=1 \ldots, T \\
& x_{t} \in X_{t}, t=1 \ldots, T 
\end{aligned}
\end{equation}

\noindent where $g(x_t,\xi_t)$ or $g_t$ is a  non-linear objective function from time $t$ to $t+1$, $h(\xi_{t}, x_{t})$ is a non-linear constraint at time $t$. The value of objective function and constraints at any time are evaluated with Monte Carlo simulations. In this paper, we solve the problem in (\ref{eq:blkstch}) to optimize the expected value of an objective function where $x_t$ is a discrete variable.

\subsection{Q-learning}
\label{Q-Learning-sec}

Reinforcement learning maximizes the cumulative expected value of the objective function in Eq. (\ref{eq:blkstch}) when the agent interacts with the environment, characterized by Markov decision process property. The agent runs the Monte Carlo simulation, given the decision $x$ at time $t$ to maximize the objective function. This simulation generates potential subsequent states in the Markov decision process. The state variable $s_t$ contains all the information about the system's status at time $t$, including all elements associated with uncertain parameters (i.e., $\xi$) at time $t$. The Markov decision process consists of $s_t$ (the state at time $t$), $x_t$ (the decision at time $t$), $g_t$ (the value of the objective function as determined by the simulation at time $t$), and $s_{t+1}$ (the forthcoming state of the system at time $t+1$, dependant on the decision and state at time $t$).

The Bellman equation forms a fundamental basis to derive the Q-learning algorithm:
\begin{equation}\label{eq:qlearning}
Q(s_t,x_t)= g_{t} + \gamma \max _{X} Q(s_{t+1}, x_t)
\end{equation}

\noindent where $\gamma$ is a parameter that discounts the maximum $Q$-value at the subsequent state $s_{t+1}$. Here, $Q$ (i.e., quality) indicates the utility of a specific decision in estimating the future value of the objective function. The function $Q(s_t,x_t)$ approximates the value of the objective function from the current time $t$ until the end of the planning phase at time $T$. The primary goal of the Q-learning algorithm is to maximize the total reward, essentially the objective function. It accomplishes this by adding the maximum reward attainable from future states which is the weighted sum of the expected values of the rewards of all future steps starting from the current state.
We iteratively approximate $Q(s_t,x_t)$ where $\alpha$ represents the learning rate, varying between 0 and 1:

\begin{equation}\label{eq:qlearningiteration}
Q(s_t,x_t) \leftarrow Q(s_t,x_t)+\alpha\left(g_{t}+\gamma \max _{X} Q(s_{t+1}, x_{t+1}) -Q(s_t,x_t) \right)
\end{equation}

Eq. (\ref{eq:qlearningiteration}) represents the fundamental concept of the Q-learning algorithm. In this research, we utilize simulation to compute the objective function and randomly determine the system's next state. In the Q-learning framework, the agent selects a decision $x$ in state $s$ based on the epsilon-greedy policy. A policy is $\epsilon$-greedy with respect to $Q$ if, for every state $s \in S$, the decision-maker selects for the greedy decision with a probability of 1-$\epsilon$ and opts a different decision randomly from the set of all decisions (both greedy and non-greedy) with a probability $\epsilon$. The value of $\epsilon$ at episode $e$ is defined as $\epsilon_e = \operatorname{max}\{\epsilon_{end}, \epsilon_{decay} \epsilon_{e-1}\}$ where $\epsilon_{decay}$ is the decay rate for $\epsilon$ (e.g., 0.999) and $\epsilon_{e-1}$ is the value of epsilon at episode $e-1$. Initially, $\epsilon$ starts near one ($\epsilon_1 =1)$, and it gradually decreases as the algorithm learns the optimal policy. A large $\epsilon$ prompts the algorithm to make random choices to explore and understand the environment, whereas a low $\epsilon$ encourages it to select the best decision (sub-optimal decision) in each state to maximize $Q$.

Algorithm \ref{alg:Qlearning} as shown in \cite{wang2013backward}, details the Q-learning procedure. After the agent repeats this process a sufficient number of times, $Q(s_t,x_t)$ converges to the optimal value (see \cite{melo2001convergence} for more detail), where:
\begin{equation}\label{eq:ss}
0 \leftarrow\left|g_{t}+\gamma \max _{X} Q\left(s_{t+1}, x_{t+1}\right)-Q\left(s_{t}, x_{t}\right)\right|
\end{equation}

\begin{algorithm}[]
	\caption{Q-learning Algorithm.}
	\label{alg:Qlearning}
	\begin{algorithmic}[1]
	    \STATE Initialize $Q$ values arbitrarily (e.g., $Q(s,x)=0$ for all $s \in S$ and $x \in X(s)$)
		\FOR{$i \leftarrow 1$ \textit{to} Total Number of Episodes ($E$)}
		    \STATE $\epsilon \leftarrow \epsilon_i$
		    \STATE Reset the simulation environment parameters
		    \STATE $t \leftarrow 1$
		    \STATE \hspace{\parindent} Select either random or optimal greedy values for decision variables $x_t$ using epsilon policy algorithm, satisfying constraints $h(\xi_{t}, x_{t})=b_{t}\left(\xi_{t}\right)$
		    \STATE \hspace{\parindent} Run simulation model to evaluate $x_t$ given uncertainty at time $t$ ($\xi$)
		    \STATE \hspace{\parindent} Evaluate objective function at time $t$ ($g_t$)
		    \STATE \hspace{\parindent} $Q(s_t,x_t) \leftarrow Q(s_t,x_t)+\alpha (g_{t}+\gamma \max_{X} Q(s_{t+1}, x_{t+1}) -Q(s_t,x_t))$
		    \STATE \hspace{\parindent}  $t \leftarrow t+1$
		    \STATE \textbf{until} $t$ is end of planning horizon
		\ENDFOR
		\STATE \textbf{return} $Q$
		\\\hrulefill
		\STATE Output: State-decision value function ($Q$) 
	\end{algorithmic}
\end{algorithm}

A $Q$-table records the $Q$-values for each ($s_t,x_t$) pair. During training, the $Q$-values are updated while the algorithm proceeds. Once the algorithm converges,  the $Q$-table shows the optimal decision for every system state. The Q-learning algorithm is guaranteed to converge to the optimal solution (see \cite{melo2001convergence}) given a small step-size $\alpha$, and every($s_t,x_t$) pair, for all $s\in S$ and $x\in X(s)$, is explored an infinite number of times. The optimal policy, which is derived using Eq. (\ref{eq:optimalpolicy}), is the one that maximizes $Q(s,x)$ for each pair. This policy represents the best decision for any possible variation of the uncertain parameter.
\begin{equation}\label{eq:optimalpolicy}
\pi^{*}(s)=\arg \max _{X} Q^{*}(s, x) \quad \forall s \in S
\end{equation}

Traditional reinforcement learning algorithms, like Q-learning, can determine the optimal policy for relatively smaller-scale problems. However, when the number of state-decision pairs increases, the Q-learning algorithm may not be able to address a sequential decision problem reliant on Monte Carlo simulations. The Deep Q-learning algorithm, an advancement over the traditional Q-learning, effectively resolves this issue.
\subsection{Deep Q-learning}
If the number of states increases to thousands, the curse of dimensionality may appear, necessitating the saving of state-decision values in a big table. It is practically impossible to solve the problem if thousands of possible states exist. Deep Q-learning algorithms can estimate $Q(s_t,x_t)$ within an acceptable error \cite{hester2017deep}. Deep Q-Learning combines Q-learning with a flexible deep neural network and has been tested on a varied and large set of Atari games, reaching human-level performance on many games \cite{van2016Deep}. Many problems are too large to learn all decision values in all states separately. Evaluating the true value for the decision-state value function is practically impossible for large problems such as a multi-stage simulation optimization problem. A Deep Q-learning algorithm estimates the decision-state value function with a Deep neural network. 
Using Deep Q-learning to solve a multi-stage stochastic programming problem where the objective function and constraints are evaluated via simulation represents a novel development in the field of engineering system design.

In multi-stage decision-making, uncertainty unfolds over time. Decisions made at time $t$ are based solely on information available at the beginning of $t$. In other words, if two scenarios share identical uncertainty data up to time $t$ , their decisions at this point should be the same, regardless of future differences \cite{cardin2017approach}. The optimal decision at any state $s$ should be independent of future uncertain parameter values.
Algorithm \ref{alg:DeepQlearning} demonstrates how the Deep Q-learning algorithm tackles the multi-stage stochastic program outlined in Eq. (\ref{eq:blkstch}). This algorithm selects decisions at each state from feasible solutions, attempting to maximize the objective function (e.g., expected profit) in a sequence of decisions and find the optimal policy for the engineering system design. The proposed framework is suitable for diverse set of objective functions evaluated via simulation with any type of constraints when uncertainty exists around the parameters given discrete decision variables.

\begin{algorithm}[]
	\caption{Deep Q-learning Algorithm for Solving Eq. (\ref{eq:blkstch}).}
	\label{alg:DeepQlearning}
	\begin{algorithmic}[1]
        \STATE Initialize neural network weights $w^- \leftarrow w$
		\FOR{$i \leftarrow 1$ \textit{to} $E$}
		    \STATE $\epsilon \leftarrow \epsilon_i$
		    \STATE $t \leftarrow 0$
		\STATE \textbf{Repeat}
		    \STATE \hspace{\parindent} \hspace{\parindent} Find feasible decision variable $x_t$ by fulfilling constraints $h(\xi_{t}, x_{t})\leq b_{t}\left(\xi_{t}\right)$ using policy $\pi \leftarrow \epsilon$-greedy$ (\hat{Q}(s_t,x_t,w))$
		    \STATE \hspace{\parindent} \hspace{\parindent} Run simulation model to evaluate 
		    $x_t$ given uncertain parameters value at time $t$
		    \STATE \hspace{\parindent} \hspace{\parindent} Find objective function at time $t$ ($g_t$) by simulation
		    \STATE \hspace{\parindent} \hspace{\parindent} Generate random sample from uncertain parameters distributions for the next period $t+1$ and update state $s_{t+1}$
		    \STATE \hspace{\parindent} \hspace{\parindent} Store experience tuple ($s_t,x_t,g_t,s_{t+1}$)
		    \STATE \hspace{\parindent} \hspace{\parindent} Obtain random minibatch of tuples ($s^j, x^j, g_t^j, s^{j+1}$) 
		    \STATE \hspace{\parindent} \hspace{\parindent} Set target $y_t^j = g_t^j + \gamma \max _{X} \hat{Q}(s_t^{j+1}, x_t, w^-)$
		    \STATE \hspace{\parindent} \hspace{\parindent} Update $\Delta w= \alpha (y_t^j - \hat{Q}(s_t^j, x_t, w)) \nabla_w \hat{Q}(s_t^j, x_t, w)$
		    \STATE \hspace{\parindent} \hspace{\parindent} Reset $w^- \leftarrow w$ every C steps
		    \STATE \textbf{until} $t$ is end of planning horizon
		\ENDFOR
		\STATE \textbf{return} $\hat{Q}$
		\\\hrulefill
		\STATE Output: State-decision value function approximation ($\hat{Q}$) 
	\end{algorithmic}
\end{algorithm}

Algorithm \ref{alg:DeepQlearning} presents a step-by-step framework for optimizing engineering system designs where the objective function is evaluated via a simulation. Initially, the algorithm assigns random values to the neural network's weights and iterates over the total number of episodes, denoted by $E$. Each episode begins by resetting the simulation environment including the state variable $s$.

Given the state of the system at time $t$, the decision variable is selected via the epsilon greedy algorithm satisfying constraints in Eq. (\ref{eq:blkstch}). This state, along with uncertain parameters $\xi$ and time $t$, serve as inputs for the deep neural network. The neural network then generates a Q-value for each potential decision derived from the decision variable ($X$). For example, if the decision variable $X$ can only take an integer value within $[0,10]$, the neural network computes an approximation of the Q-value for each feasible decision. It estimates $Q(s_t, x_t)$ for each decision at every state. The agent engages with the simulation environment by drawing samples from the uncertain parameters. 
A random sample becomes an input for the deep neural network, and the optimal decision is calculated based on the $\epsilon$-greedy policy.

As a function approximator, the deep neural network takes the system's state and outputs a vector of decision values, with the highest value indicating the optimal decision. Initially, when the deep neural network starts with random values, the solutions tend to be random and sparse, often leading to suboptimal choices. Over time, the Deep Q-learning algorithm identifies and learns the most effective decisions for each state. This differs from traditional reinforcement learning algorithms like Q-learning, which produce only one Q-value at a time. In contrast, Deep Q-Learning generates a Q-value for every possible decision in a single forward pass.

After the objective function is calculated via a simulation, the system's next state, $s_{t+1}$ is determined by sampling from the distribution of the random variable, $\xi$. Decisions are then made based on the deep neural network's output, using the $\xi$-greedy algorithm satisfying constraints. The neural network's weights are updated using the gradient descent algorithm.

\begin{figure}[H]
	\centering
	\includegraphics[trim={0cm 19cm 0cm 0cm},clip,width=9cm]{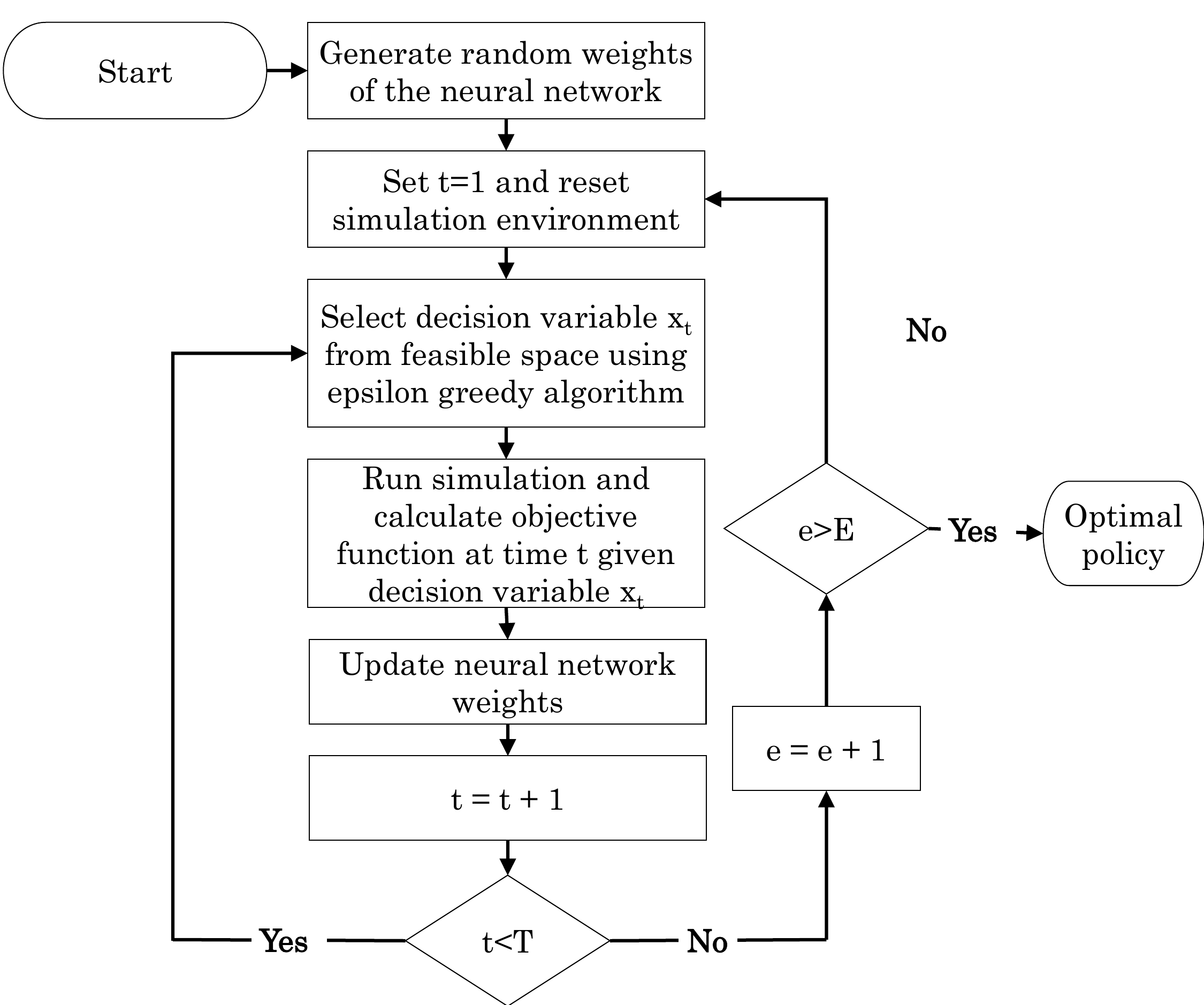}
	\captionsetup{justification=centering}
	\caption{Proposed framework for solving multi-stage stochastic problem}
	\label{fig:framework}
\end{figure}

Upon reaching the planning horizon's end, the algorithm resets the simulation environment and starts a new episode. This iteration continues until the agent fully learns the environment and discovers the optimal policy to maximize the objective function, which could be profit or another simulation output. At the end of training, the optimal policy $\pi^*$ is identified. The optimal policy is a trained neural network ($\hat{Q}$). This policy delivers the best decision for given inputs, including the state of the system and uncertain parameters. The algorithm is also illustrated in Figure \ref{fig:framework}. The objective function in (\ref{eq:blkstch}) is approximated by simulating the optimal policy (trained neural network) $M$ times, randomly sampling $\xi$ and utilizing policy $\pi^*$:

\begin{equation}\label{eq:apppolicy}
\begin{aligned}
& \max f(x, \xi) = \frac{1}{K}\sum\limits_{m=1}^{M}{\sum\limits_{t=1}^{T}{\hat{Q}_{\pi^*}({x_t}^*,\xi_m)}} 
\end{aligned}
\end{equation}

\section{Engineering System Design Illustrative Examples}

The framework outlined in this paper can be applied to dynamic stochastic programming problems with a discrete number of decisions at each stage. The illustrative examples presented in this section serve as a proof-of-concept, showcasing the framework's capabilities. The illustrative examples aim to illustrate how the algorithm functions. The results indicate that the Deep Q-learning algorithm is effectively applicable to engineering design problems with a sequential decision-making process under uncertainty, where the objective function is evaluated through simulation. 
\subsection{Capacity Expansion Problem with Price Uncertainty: Problem Formulation}\label{Trnprc}

The illustrative example is a capacity expansion problem under uncertainty. Decision makers, such as those involved in power plant design \cite{murphy2005generation,wogrin2011generation,giahi2022optimizing} or transportation network design \cite{mathew2009capacity}, face the challenge of deciding when to expand operational capacity. In the context of power plants, this might involve adding elements like wind turbines. The designers would like to plan for when they should expand the capacity of their operations, but uncertainty in the future makes the decision of whether to expand their capacity challenging. The designers may choose to expand their capacity under some scenarios and choose not to expand their capacity under other scenarios. 

The decision maker's goal is to find the optimal policy to design a system operating over $T$ time periods, structured as a multi-stage stochastic program. The decision involves determining $x_t$, the additional capacity units to add at time $t$\ to maximize $f(x,\xi)$, the expected profit of the system. This profit is calculated as expected revenue minus the cost of expansion and operation over the $T$ periods. 
The revenue is uncertain and is a function of $\xi$, the future price $p_t$ gained from selling one unit of capacity per time unit. Here, revenue at time $t$ is $up_t\sum_{t'=1}^tx_{t'}$, where $u$ represents the number of units produced.
While $p_t$ is known when the decision-maker determines $x_t$, future prices $p_{t+1}$ remain uncertain and depend on the current price $p_t$. The problem is defined as follows: 

\begin{equation}\label{eq:example1}
\begin{aligned}
 \max f(x, \xi) & = \left(up_1-c_{om}-c_{inv}\right)x_1+\\&E\left[ \sum_{t=2}^T \dfrac{\left(up_t-c_{om}\right)\sum_{t'=1}^tx_{t'}-c_{inv}x_t}{\left(1+i\right)^{t-1}}\right]\\
 \text{subject to } & p_{t+1} =  p_t e^{N(\mu_1, \sigma_1)}, \; \; t=1, \ldots, T-1\\
 & \sum_{t=1}^Tx_t\leq 1\\
 & x_t \in \{0,1\}, \; \; t=1, \ldots, T\\
 %
\end{aligned}
\end{equation}
%
%
%
%
%
%
%
%

\noindent where $c_{om}$ is the annual operating cost, $c_{inv}$ refers to the cost of adding capacity, $i$ represents interest rate, and $\mu_1$ and $\sigma_1$ represent the mean and standard deviation to describe the random evolution of the price.

The optimization problem in (\ref{eq:example1}), similar to the multi-stage stochastic program in (\ref{eq:blkstch}), solves for optimal solutions at each planning period for all potential $p_t$ values, for $t=1,2,\ldots,T$.
A key constraint is the total capacity not exceeding 1, indicating that capacity addition can only occur in one stage. This simplified model emphasizes the timing and feasibility of building capacity. The system has no initial capacity. 
At time $t$, the system gains revenue by selling accumulative capacity (i.e., $ \sum\limits_{t^{\prime}=1}^{t} x_{t^{\prime}}$) at price $p_t$, where consecutive prices ratios follow a lognormal distribution with $\mu_1$ and $\sigma_1$ parameters.
The operations cost, $c_{om}$ is the cost to operate the total capacity existing at time $t$, and investment cost, $c_{inv}$ applies if capacity is added at time $t$ (i.e., $x_t$).     

\subsection{Capacity Expansion Problem with Price Uncertainty: Solution}

We first define the state ($S$), decision variable ($X$), and the objective function in detail. The state $S$ encompasses two components: the current price $p_t$ and the residual capacity at time $t$, represented as $1-\sum_{t'=1}^tx_{t'}$. 
The decision revolves around whether to add capacity at each stage of the decision-making process. This decision is a binary variable, and the algorithm makes a decision at time $t$ given state $s$. At each step, the model accrues a reward $g(p_t, x_t)=\left[\left(up_t-c_{om}\right)\sum_{t'=1}^tx_{t'}-c_{inv}x_t\right]/\left(1+i\right)^{t-1}$ after executing decision $x_t$ and its subsequent evaluation via simulations. The objective is to maximize expected profits while satisfying the constraints. 

Given that $p_t$ is a continuous random variable, we select $10,000$ samples from the price distribution for each time period. In this problem, $c_{om} = 300$, $c_{inv} = 20$, $u = 2920$, $\mu_1 = 0.05$, $\sigma_1 = 0.1$, and $p_1 = 0.1$. The Deep-Q Network ($DQN$) algorithm runs for 150,000 episodes (iterations). The $\epsilon_e$ parameter is calculated as $max(\epsilon_{end}, \epsilon_{decay} \times \epsilon_{t-1})$, where $\epsilon_1 = 1$, $\epsilon_{end}= 0.001$, and $\epsilon_{decay} = 0.99995$. 



\begin{figure}
     \centering
     \includegraphics[width=\linewidth]{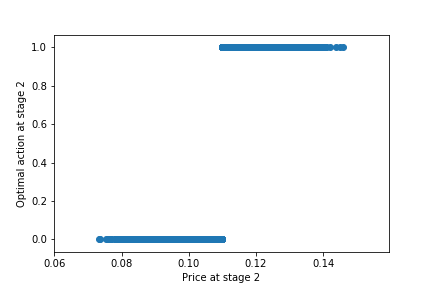}
     \caption{Optimal decision at stage two for two-stage problem with price uncertainty}
     \label{fig:decision_p22}
\end{figure}

This methodology is applied to two case studies from the capacity expansion problem, as mentioned in (\ref{eq:example1}). In the first case, with $T=2$ stages, 
the decision maker should select $x_1=0$ at time $t=1$. The optimal decision at time $t=2$ depends on the price: if $p_2\geq 0.1096$, then $x_2=1$ is the optimal decision; otherwise if $p_2< 0.1096$, the optimal decision is $x_2=0$. This simple problem can be solved analytically, indicating $x_2=1$ if $x_1=0$ and $up_2 - c_{om} - c_{inv}>0$, when $p_2\geq 0.1096$. The algorithm's threshold for $p_2$ when $x_2=1$ is with 0.0001 of the analytical solution. The simulated expected profit for this policy is 7.38, and the algorithm converges the optimal solution in 980 seconds. Figure \ref{fig:decision_p22} depicts the optimal decision at $t=2$ given $p_2$.

\begin{figure}[t]
     \centering
         \includegraphics[width=\linewidth]{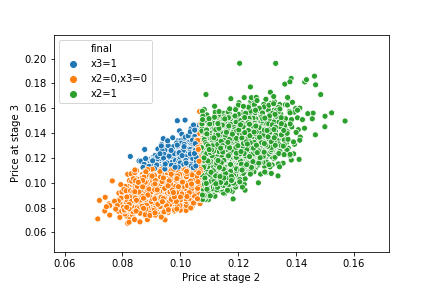}
         \caption{Optimal decision at stages 2 and 3 for three-stages problem with price uncertainty}
         \label{fig:p32_a3}
\end{figure}

In the second case, with $T=3$ stages, the optimal decision begins similarly with 
$x_1=0$ at time $t=1$. 
The algorithm converges to the optimal solution after approximately 80,000 episodes, taking 1350 seconds. As shown in Figure \ref{fig:p32_a3}, the optimal policy at stages 2 and 3 depends on $p_t$. The algorithm suggests adding a capacity at $t=2$ if $p_2 \geq 0.1072$. If $p_2 < 0.1072$, the algorithm determines that $x_2=0$ and $x_3=1$ if $p_3 > 0.1110$. This approach is derived from system simulations and random price sampling from the price distribution, with decisions optimized for each given state using the optimal policy, as estimated in Eq. (\ref{eq:apppolicy}). The expected profit here is 32.5. Analytically, the decision-maker should add one capacity at stage 3 if $p_3 > 0.1096$ and have no increase in stage 2. The decision maker should add capacity in stage 2 if

\begin{equation}
\begin{aligned}
&up_2-c_{om}-c_{inv}+\dfrac{u\operatorname{E}\left[p_3\vert p_2\right]-c_{om}}{1+i}> \\
&P\left(p_3>\frac{c_{om}+c_{inv}}{u} \Bigg\vert p_2\right)\dfrac{u\operatorname{E}\left[p_3\vert p_2,\;p_3> \frac{c_{om}+c_{inv}}{u}\right]-c_{om}-c_{inv}}{1+i}.
\end{aligned}
\end{equation}

Monte Carlo simulation indicates capacity addition at stage 2 if $p_2 > 0.1061$. The Deep-Q learning algorithm's thresholds for $p_2$ and $p_3$ are remarkably close to the true thresholds, differing by 0.0011 and 0.0014, respectively. The expected profits between the two solutions vary slightly, by about 0.08 or 0.2\%, which is considered relatively insignificant.

\subsection{Capacity Expansion Problem with Price and Demand Uncertainty: Problem Formulation}\label{Trnprcdmd}

The Deep Q-learning algorithm can handle multiple sources of uncertainty. The uncertain parameters are stored as a tuple of state variable parameters and passed to the policy model as input. The neural network approximates the relationship between the inputs (state) and the outputs (decision) while new stream of data---produced by simulation---comes to the model. The reinforcement learning algorithm enables the decision-maker to explore the simulation environment and learn the optimal policy.

We add demand as another source of uncertainty in this illustrative capacity expansion example. In many engineering problems, demand is an uncertain variable that can significantly impact how a system is constructed and how much capacity is required. In this example, the decision maker can decide to add a total of $K$ units of capacity to the system. If the demand is high, then multiple units of capacity may be needed. The decision maker will decide when and under what conditions to add a new capacity. The decision maker's goal is to maximize the expected profit but the system will only receive revenue for what is demanded, and the revenue at time $t$ is $up_t\times\min\{d_t,c_p\sum_{t'=1}^tx_{t'}\}$ where $d_t$ is the demand at time $t$ and $c_p$ is the units of demand that can be satisfied for each unit of capacity. When the decision maker determines $x_t$, the decision maker knows $p_t$ and $d_t$, but the future price $p_{t+1}$ and demand $d_{t+1}$ are uncertain and depend on the price and demand at time $t$. The problem is defined as follows: 

\begin{equation}\label{eq:example2}
\begin{aligned}
 \max f(x, \xi) & =  up_1\times\min\{d_1,c_px_{1}\}-(c_{om}-c_{inv})x_1+ \\ 
 & E\left[ \sum_{t=2}^T \dfrac{up_t\times\min\{d_t,c_p\sum_{t'=1}^tx_{t'}\}-c_{om}\sum_{t'=1}^tx_{t'}-c_{inv}x_t}{\left(1+i\right)^{t-1}}\right]\\
 \text{subject to } & p_{t+1} =  p_t e^{N(\mu_1, \sigma_1)}, \; \; t=1, \ldots, T-1\\
 & d_{t+1} =  d_t e^{N(\mu_2, \sigma_2)}, \; \; t=1, \ldots, T-1\\
 & \sum\limits_{t=1}^{T} x_t \leq K \\
 & x_t \in \{0, \ldots,K\}, \; \; t=1, \ldots, T\\
 %
\end{aligned}
\end{equation}

\subsection{Capacity Expansion Problem with Price and Demand Uncertainty: Solution}

To evaluate the performance of the proposed framework on the capacity expansion problem with multiple sources of uncertainty, we run the algorithm to find the optimal policy for the problem described in Eq. (\ref{eq:example2}). $\xi_t$ is a tuple of $p_t$ and $d_t$ which makes the stochastic problem more complex than the problem with only one source of uncertainty. Since $p_t$ and $d_t$ are continuous random variables, the algorithm selects 100,000 samples from the distribution of the price and demand at each time period. In this example, $d_1=1$, $\mu_2= 0.2$, $\sigma_2 = 0.1$, and $c_p= 1$. 

\begin{figure*}[t!]
     \centering
     \begin{subfigure}[b]{0.45\linewidth}
         \centering
         \includegraphics[width=\linewidth]{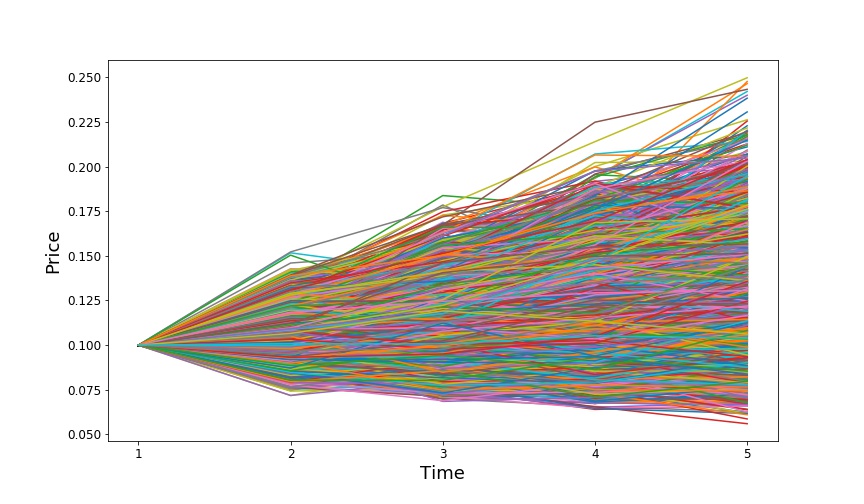}
         \caption{Price Uncertainty}
         \label{fig:price}
     \end{subfigure}
     \begin{subfigure}[b]{0.45\textwidth}
         \centering
         \includegraphics[width=\linewidth]{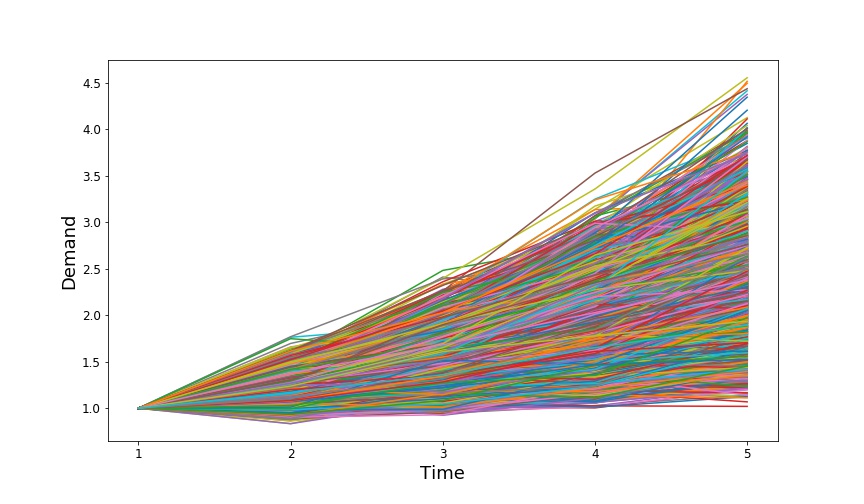}
         \caption{Demand Uncertainty}
         \label{fig:demand}
     \end{subfigure}
    \caption{Distribution of uncertain parameters during the planning horizon}     
\end{figure*}

Figure \ref{fig:price} and \ref{fig:demand} show price and demand, respectively. These figures show that the demand and price are highly uncertain and the combination of these two sources of uncertainty creates many scenarios.

We use the proposed framework to solve the problem for $T=3,4,$ and 5 stages. Table \ref{tab:policy_price_dem} shows the optimal expected profit when the number of stages $T$ varies between 3 and 5 and maximum capacity ($K$) varies between 2 and 4. In addition, Table \ref{tab:policy_price_dem}  shows the time it takes to run each of the configurations.

\begin{table}[t]
\centering
\caption{Expected profit of multi-stage stochastic problem under demand and price uncertainty}
\begin{tabular}{|l|l|l|l|}
\hline
(Expected Profit,\\ Run Time (sec)) & \multicolumn{3}{c|}{Number of Stages}    \\ \hline
K                                     & 3           & 4           & 5            \\ \hline
2                                     & (33.1,1411) & (77.8,1650) & (135.3,2052) \\ \hline
3                                     & -           & (80.1,1753) & (146.5,2109) \\ \hline
4                                     & -           & -           & (150.8,2172) \\ \hline
\end{tabular}
\label{tab:policy_price_dem}
\end{table}

\begin{figure*}][t]
     \centering
     \begin{subfigure}[b]{0.2\textwidth}
         \centering
         \includegraphics[width=\linewidth]{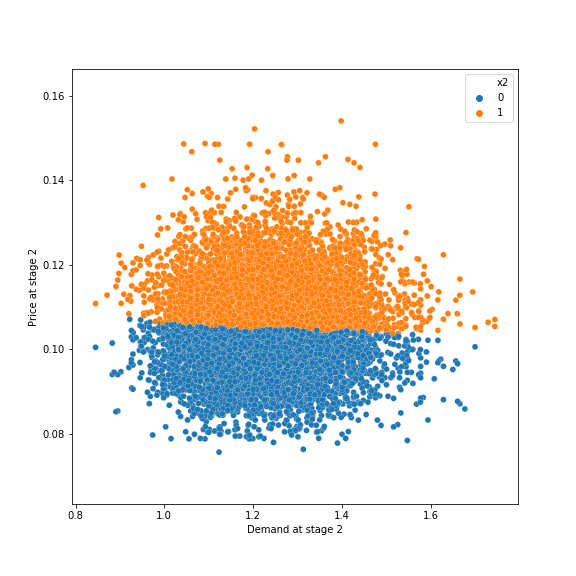}
         \caption{Optimal decision at stage 2}
         \label{fig:s3k2-x2}
     \end{subfigure}
    \hfill
     \begin{subfigure}[b]{0.2\textwidth}
         \centering
         \includegraphics[width=\linewidth]{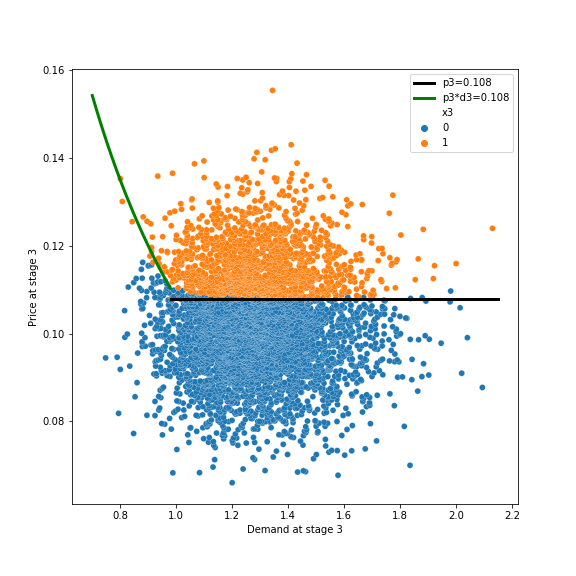}
         \caption{Optimal decision at stage 3 given $x_2=0$}
         \label{fig:s3k2-x30}
     \end{subfigure}
    \hfill
     \begin{subfigure}[b]{0.2\textwidth}
         \centering
         \includegraphics[width=\linewidth]{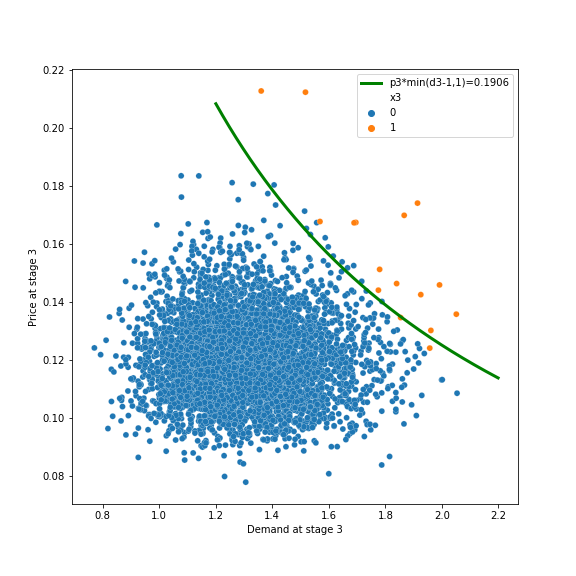}
         \caption{Optimal decision at stage 3 given $x_2=1$}
         \label{fig:s3k2-x31}
     \end{subfigure}
    \hfill
     \begin{subfigure}[b]{0.2\textwidth}
         \centering
         \includegraphics[trim={0cm 0cm 0cm 1.12cm},clip,width=\linewidth]{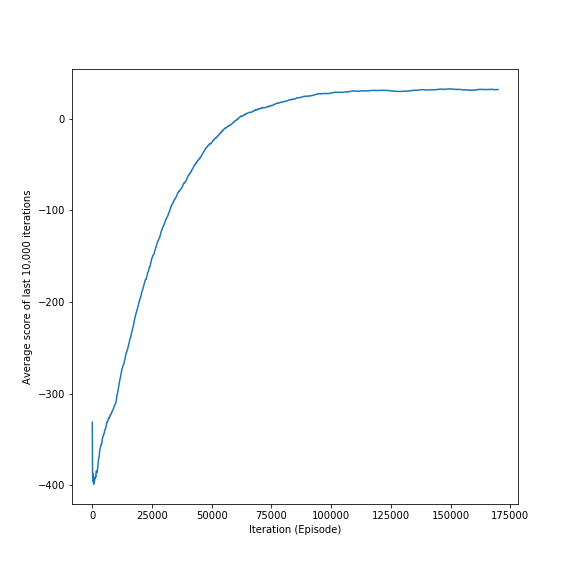}
         \caption{Algorithm convergence to the optimal policy}
         \label{fig:convergencepd33}
     \end{subfigure}
    \caption{Three-stage problem with maximum capacity of two}
    \label{fig:pd33all}

\end{figure*}

When $T=3$ and $K=2$, the algorithm converges to a solution after 100,000 episodes as seen in Figure \ref{fig:convergencepd33}. The expected profit is 33.1. As with the previous example, the decision maker should choose $x_1=0$.

Since an analytical solution can use backward dynamic programming techniques, the analysis of the solution will begin with the decision in the last stage ($t=3$) and proceed to the decision at $t=2$.
An analytical solution reveals that if $x_2=0$, the decision maker should choose $x_3=1$ if $up_3\times\operatorname{min}\{d_3,1\}-c_{om}-c_{inv}>0$ and choose $x_3=2$ if $up_3\times\operatorname{min}\{d_3,2\}-2\left(c_{om}-c_{inv}\right)>up_3-c_{om}-c_{inv}>0$. If $x_2=1$, the decision maker should choose $x_3=1$ if $up_3\times\operatorname{min}\{d_3-1,1\}-c_{om}-c_{inv}>0$.
Given the numbers in this example, if $x_2=0$, then $x_3=2$ if $p_3\times\operatorname{min}\{d_3-1,1\}>0.1906$ and $x_3=1$ if $p_3\times\operatorname{min}\{d_3,1\}>0.1096$ and the condition for $x_3=2$ is not met. Figure \ref{fig:s3k2-x30} depicts the algorithm's solution for $x_3$ if $x_2=0$. The algorithm finds a threshold for $p_3$ that separates $x_3=0$ and $x_3=1$ when $d_3\geq 1$. This threshold occurs at approximately $p_3=0.108$, which is within 0.0016 of the analytical threshold. When $d_3<1$, the dividing line between $x_3=0$ and $x_3=1$ approximately follows $p_3d_3=0.1906$. The Deep-Q learning algorithm never finds $x_3=2$ as an optimal policy perhaps because only a few instances exist when $x_3=2$ should be optimal. If $x_2=1$, then $x_3=1$ if $p_3\times\operatorname{min}\{d_3-1,1\}>0.1906$. 
Figure \ref{fig:s3k2-x31} displays the algorithm's solution for $x_3$ if $x_2=1$. The algorithm finds a dividing line between $x_3=0$ and $x_3=1$ that approximately follows $p_3\operatorname{min}\{d_3-1,1\}=0.1906$ when $x_2=1$. 

Figure \ref{fig:s3k2-x2} shows the algorithm's solution for $x_2$. The algorithm finds a threshold that separates $x_2=0$ and $x_2=1$ if $d_2>1.2$ and $p_2=0.104$ If $d_3<1.2$, the threshold separating $x_2=0$ and $x_2=1$ appears to follow $p_2d_2=0.077$.


\begin{figure*}[!t]
     \centering
     \begin{subfigure}[b]{0.2\linewidth}
         \centering
         \includegraphics[width=\linewidth]{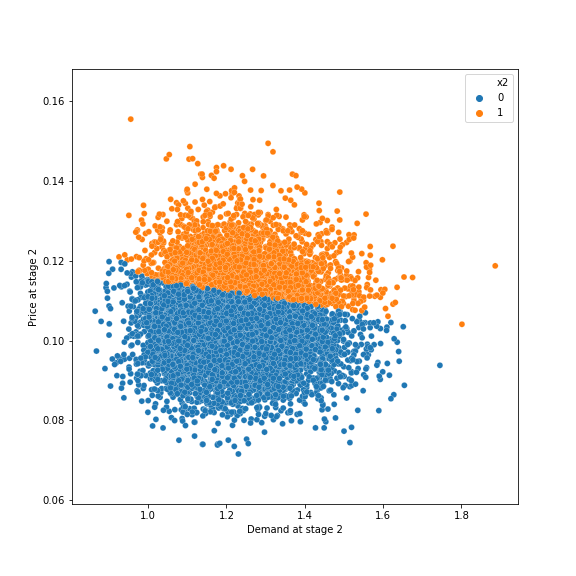}
         \caption{optimal decision at stage 2}
         \label{fig:s4k3-x2}
     \end{subfigure}
     \hfill
     \begin{subfigure}[b]{0.2\textwidth}
         \centering
         \includegraphics[width=\linewidth]{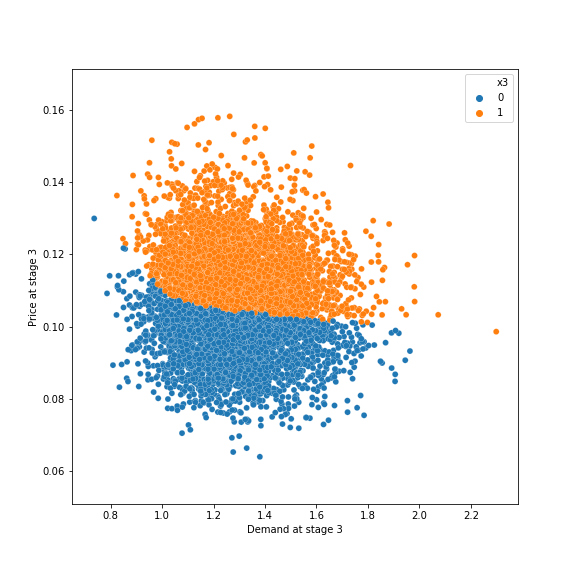}
         \caption{optimal decision at stage 3 given $x_2=0$}
         \label{fig:s4k3-x30}
     \end{subfigure}
     \hspace{0\baselineskip}
     \hfill
     \begin{subfigure}[b]{0.2\textwidth}
         \centering
         \includegraphics[width=\linewidth]{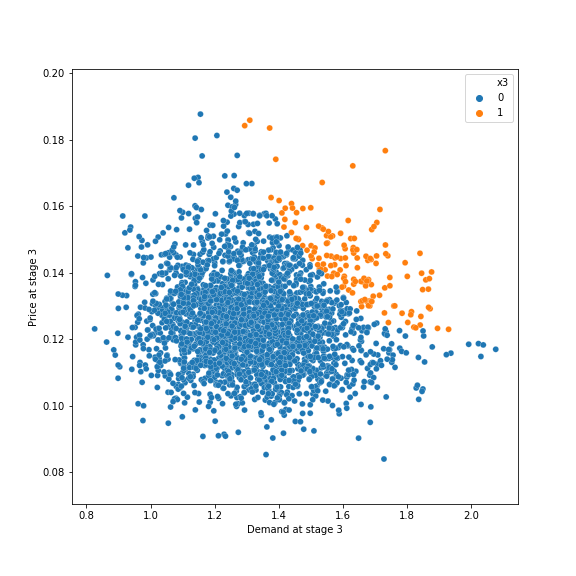}
         \caption{optimal decision at stage 3 given $x_2=1$}
         \label{fig:s4k3-x31}
     \end{subfigure}
     \hfill
     \begin{subfigure}[b]{0.2\textwidth}
         \centering
         \includegraphics[width=\linewidth]{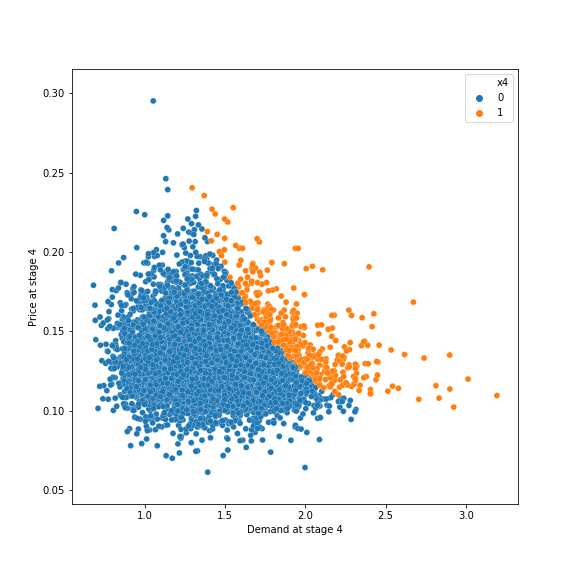}
         \caption{optimal decision at stage 4 given $x_2+x_3=1$}
         \label{fig:s4k3-x2+x3=1}
     \end{subfigure}
     \hspace{0\baselineskip}
     \hfill
     \begin{subfigure}[b]{0.2\textwidth}
         \centering
         \includegraphics[width=\linewidth]{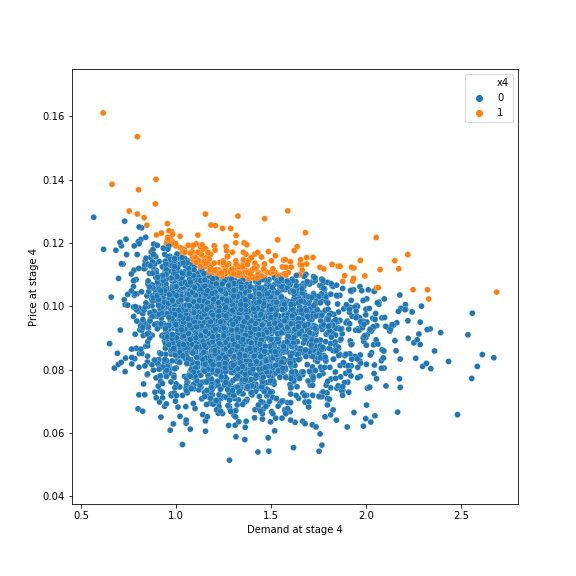}
         \caption{optimal decision at stage 4 given $x2=0,x3=0$}
         \label{fig:s4k3-x2=0x3=0}
     \end{subfigure}
     \hspace{0\baselineskip}
     \begin{subfigure}[b]{0.2\textwidth}
         \centering
         \includegraphics[width=\linewidth]{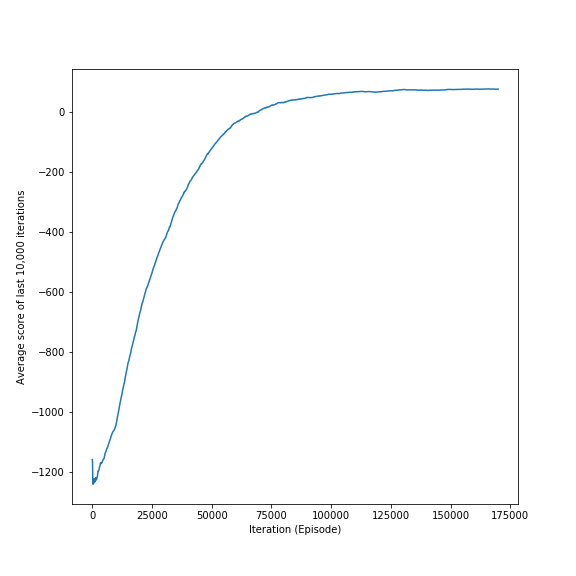}
         \caption{Algorithm convergence to the optimal policy}
         \label{fig:convergencepd44}
     \end{subfigure}
     \hspace{0\baselineskip}
    \caption{Four-stage problem with maximum capacity of three}
    \label{fig:pd44all}
\end{figure*}

If $T=4$ and $K=3$, the algorithm finds a solution with an expected profit of 80.1. If $x_2=x_3=0$, the final decision $x_4=1$ if $d_4<1$ and $p_4d_4> 0.1906$, or if $d_4\geq 1$ and $p_4>0.1906$. If $x_2=x_3=0$, the final decision $x_4=2$ if $p_4\times \operatorname{min}\{d_4-1,1\}>0.1906$. As shown in Figure \ref{fig:s4k3-x2=0x3=0}, the Deep-Q learning algorithm identifies $p_4=0.1906$ as a dividing line between $x_4=0$ and $x_4=1$, but the line begins at approximately $d_4=1.2$ rather than at $d_4=1$. The algorithm does not find any scenarios where $x_4=2$. If either $x_2=1$ or $x_3=1$, then $x_4=1$ if $p_4\times\operatorname{min}\{d_4-1,1\}>0.1906$. Figure \ref{fig:s4k3-x2+x3=1} shows that the Deep-Q learning algorithm's solution approaches the analytical solution although the slope of the dividing line between $x_4=0$ and $x_4=1$ is less steep than the true solution when $d_4<2$.

Figure \ref{fig:s4k3-x30} depicts the algorithm's solution for choosing $x_3=0$ or $x_3=1$ when $x_2=0$. The algorithm identifies a clear threshold, and the slope of this dividing line is steep when $d_3<1.15$ and much less steep when $d_3>1.3$. The algorithm does not identify any situations when $x_3\geq 2$ or when $x_3=2$ given $x_2=1$. As seen in Figure \ref{fig:s4k3-x2}, a fairly clear threshold exists for the decision at $t=2$.


\section{Conclusion}

This paper presents a framework for identifying the optimal policy in engineering system design. When designing engineering systems for long-term operation, considering uncertain future scenarios is crucial in decision-making. Our framework efficiently determines the optimal design for engineering systems under multiple uncertainties. This methodology streamlines the design process by simulating optimal designs instead of solving numerous optimization problems. The goal is to find policies that maximize the output of a simulation model given multiple sources of uncertainties. The algorithm handles both linear and non-linear multi-stage stochastic problems, where decision variables are discrete and both the objective function and constraints are evaluated via Monte Carlo simulation. 

The algorithm's effectiveness is demonstrated through two engineering design examples. The first example calculates the optimal policy for a problem with a single uncertainty source and a linear objective function. The second example calculates the optimal policy for a problem with two sources of uncertainties and a non-linear objective function. These examples highlight the methodology's ability to identify the optimal policy under various uncertain conditions, requiring fewer learning episodes compared to the number of scenarios. Both problems utilize Monte Carlo simulation to evaluate the objective function and constraints. 

This study's findings must be considered alongside its limitations. While the algorithm successfully identifies optimal policies, there is a minor error margin in the expected profit. For instance, in the first example, the expected profits from analytical and Deep Q-learning solutions differ by approximately 0.2\%. Another limitation is the framework's current inability to tackle multi-stage stochastic problems with continuous decision variables. Future research could explore advanced reinforcement learning techniques like actor-critic methods for such problems. 
Although the examples used are simplistic, they effectively demonstrate the application of the Deep-Q learning algorithm in engineering design contexts. Future research could extend this framework to more complex, real-world engineering design problems, offering deeper insights into its advantages and limitations. However, this framework significantly benefits decision-making in engineering system design, enabling decision-makers to identify optimal policies for multi-stage stochastic problems through simulation, avoiding the need to solve numerous optimization problems. For instance, the algorithm can resolve a four-stage problem with $10,000$ potential scenarios using only 150,000 iterations, in contrast to solving $10,000^4$ optimization problems. Ultimately, the optimal policy derived from the algorithm helps engineering system designers to make informed decisions under any possible future uncertainty.

\bibliographystyle{IEEEtran}
\balance
\bibliography{DQN.bib} 

\end{document}